\documentclass[11pt,a4paper]{article}
\usepackage[hyperref]{emnlp2020}
\usepackage{times}
\usepackage{latexsym}

\usepackage{rotating}
\usepackage{blindtext}
\usepackage{amsmath}
\usepackage{amsfonts}
\usepackage{graphicx}
\usepackage{mathtools}
\usepackage{booktabs}
\usepackage{relsize}
\usepackage{float}
\usepackage{graphics}
\usepackage{multirow}
\usepackage{algorithm}
\usepackage{algpseudocode}
\usepackage[utf8]{inputenc}

\DeclareUnicodeCharacter{01B0}{\`u}
\DeclareUnicodeCharacter{01A1}{\`o}

\usepackage{threeparttable}
\usepackage{booktabs, caption, makecell}

\usepackage[title]{appendix}
\usepackage[cal=cm]{mathalfa}
\usepackage[colorinlistoftodos]{todonotes}
\usepackage{multirow}
\usepackage[export]{adjustbox}
\usepackage{subcaption}
\newcommand{\remove}[1]{}
\newcommand{\mycomment}[1]{}                     
\newcommand{\margcomment}[1]{}                     

\newcommand{\ignore}[1]{}

\usepackage{microtype}
\usepackage[subtle]{savetrees}
\aclfinalcopy 

\title{{G}raph-to-{G}raph {T}ransformer for {T}ransition-based {D}ependency {P}arsing}

\author{Alireza Mohammadshahi \\ 
        Idiap Research Institute and EPFL / Switzerland 
        \AND 
        James Henderson\\
         Idiap Research Institute / Switzerland\\
  \texttt{\{alireza.mohammadshahi, james.henderson\}@idiap.ch} \\ }
  
\begin{document}
\maketitle

\begin{abstract}
  We propose the Graph2Graph Transformer architecture for conditioning on and predicting arbitrary graphs, and apply it to the challenging task of transition-based dependency parsing.  After proposing two novel Transformer models of transition-based dependency parsing as strong baselines, we show that adding the proposed mechanisms for conditioning on and predicting graphs of Graph2Graph Transformer results in significant improvements, both with and without BERT pre-training.
  The novel baselines and their integration with Graph2Graph Transformer significantly outperform the state-of-the-art in traditional transition-based dependency parsing on both English Penn Treebank, and 13 languages of Universal Dependencies Treebanks.  Graph2Graph Transformer can be integrated with many previous structured prediction methods, making it easy to apply to a wide range of NLP tasks.
\end{abstract}

\section{Introduction}

In recent years, there has been a huge amount of research on applying self-attention models to NLP tasks. Transformer~\cite{vaswani2017attention} is the most common architecture, which can capture long-range dependencies by using a self-attention mechanism over a set of vectors.  To encode the sequential structure of sentences, typically absolute position embeddings are input to each vector in the set, but recently a mechanism has been proposed for inputting relative positions \cite{shaw-etal-2018-self}.  For each pair of vectors, an embedding for their relative position is input to the self-attention function.  This mechanism can be generalised to input arbitrary graphs of relations.

We propose a version of the Transformer architecture which combines this attention-based mechanism for conditioning on graphs with an attention-like mechanism for predicting graphs and demonstrate its effectiveness on syntactic dependency parsing.  We call this architecture Graph2Graph Transformer.
This mechanism for conditioning on graphs differs from previous proposals in that it inputs graph relations as continuous embeddings, instead of discrete model structure~(e.g.\ \cite{henderson-2003-inducing,henderson-etal-2013-multilingual,dyer-etal-2015-transition}) or predefined discrete attention heads~(e.g.\ \cite{ji-etal-2019-graph,strubell-etal-2018-linguistically}).
An explicit representation of binary relations is supported by inputting these relation embeddings to the attention functions, which are applied to every pair of tokens.
In this way, each attention head can easily learn to attend only to tokens in a given relation, but it can also learn other structures in combination with other inputs.  This gives a bias towards attention weights which respect locality in the input graph but does not hard-code any specific attention weights.

We focus our investigation on this novel graph input method and therefore limit our investigation to models which predict the output graph one edge at a time, in an auto-regressive fashion.  In auto-regressive structured prediction, after each edge of the graph has been predicted, the model must condition on the partially specified graph to predict the next edge of the graph.  Thus, our proposed Graph2Graph Transformer parser is a transition-based dependency parser.  At each step, the model predicts the next parsing decision, and thereby the next dependency relation, by conditioning on the partial parse structure specified by the previous decisions.
It inputs embeddings for the previously specified dependency relations into the Graph2Graph Transformer model via the self-attention mechanism.  
It predicts the next dependency relation using only the vectors for the tokens involved in that relation.

To evaluate this architecture, we also propose two novel Transformer models of transition-based dependency parsing, called Sentence Transformer, and State Transformer. Sentence Transformer computes contextualised embeddings for each token of the input sentence and then uses the current parser state to identify which tokens could be involved in the next valid parse transition and uses their contextualised embeddings to choose the best transition. For State Transformer, we directly use the current parser state as the input to the model, along with an encoding of the partially constructed parse graph, and choose the best transition using the embeddings of the tokens involved in that transition.  Both baseline models achieve competitive or better results than previous state-of-the-art traditional transition-based models, but we still get substantial improvement by integrating Graph2Graph Transformer with them. 

We also demonstrate that, despite the modified input mechanisms, this Graph2Graph Transformer architecture can be effectively initialised with standard pre-trained Transformer models.  Initialising the Graph2Graph Transformer parser with pre-trained BERT~\cite{bert:18} parameters leads to substantial improvements. The resulting model significantly improves over the state-of-the-art in traditional transition-based dependency parsing.

This success demonstrates the effectiveness of Graph2Graph Transformers for conditioning on and predicting graph relations.  This architecture can be easily applied to other NLP tasks that have any graph as the input and need to predict a graph over the same set of nodes as output.

In summary, our contributions are:
\vspace{-2ex}
\begin{itemize}\addtolength{\itemsep}{-2ex}
\item We propose Graph2Graph Transformer for conditioning on and predicting graphs.
\item We propose two novel Transformer models of transition-based dependency parsing.
\item We successfully integrate pre-trained BERT initialisation in Graph2Graph Transformer.
\item We improve state-of-the-art accuracies for traditional transition-based dependency parsing.\footnote{Our implementation is available at: \url{https://github.com/alirezamshi/G2GTr}}
\end{itemize}

\section{Transition-based Dependency Parsing}
\label{ref:transition}
Our transition-based parser uses arc-standard parsing sequences \citep{nivre-2004-incrementality}, which makes parsing decisions in bottom-up order.  The main data structures for representing the state of an arc-standard parser are a buffer of words and a stack of partially constructed syntactic sub-trees.  At each step, the parser chooses between adding a leftward or rightward labelled arc between the top two words on the stack ({\tt LEFT-ARC($l$)} or {\tt RIGHT-ARC($l$)}, where $l$ is a dependency label) or shifting a word from the buffer onto the stack ({\tt SHIFT}).
To handle non-projective dependency trees, we allow the {\tt SWAP} action proposed in \newcite{nivre-2009-non}, which shifts the second-from-top element of the stack to the front of the buffer, resulting in the reordering of the top two elements of the stack.

\section{Graph2Graph Transformer}

\label{sec:model}

\begin{figure*}[!ht]
	\centering
	\begin{subfigure}{0.55\textwidth} 
		\includegraphics[width=\textwidth]{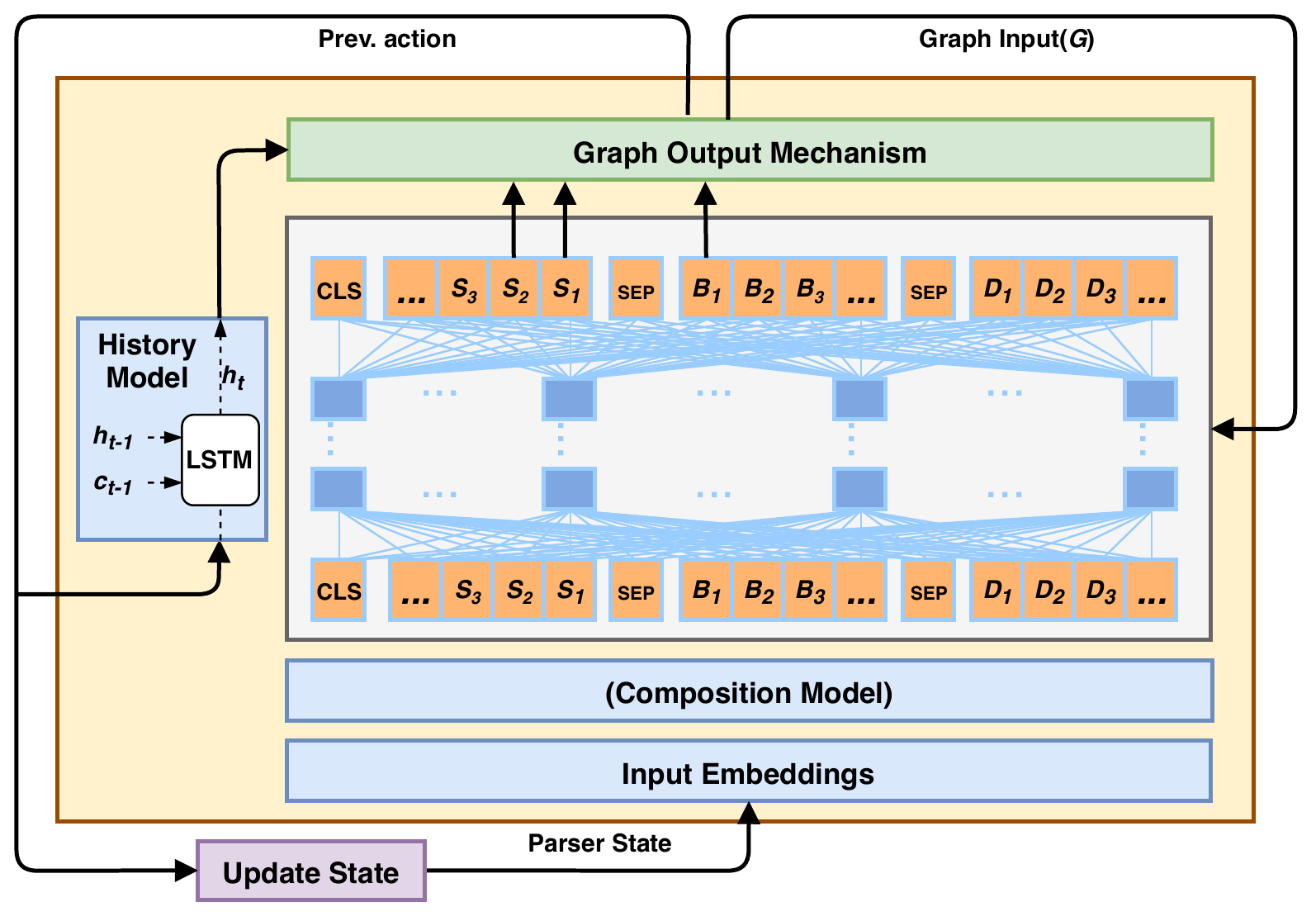}
                \vspace{-3ex}
		\caption{StateTr+G2GTr.\label{statetr}} 
	\end{subfigure}
	\hspace{1em} 
	\begin{subfigure}{0.35\textwidth} 
		\includegraphics[width=\textwidth]{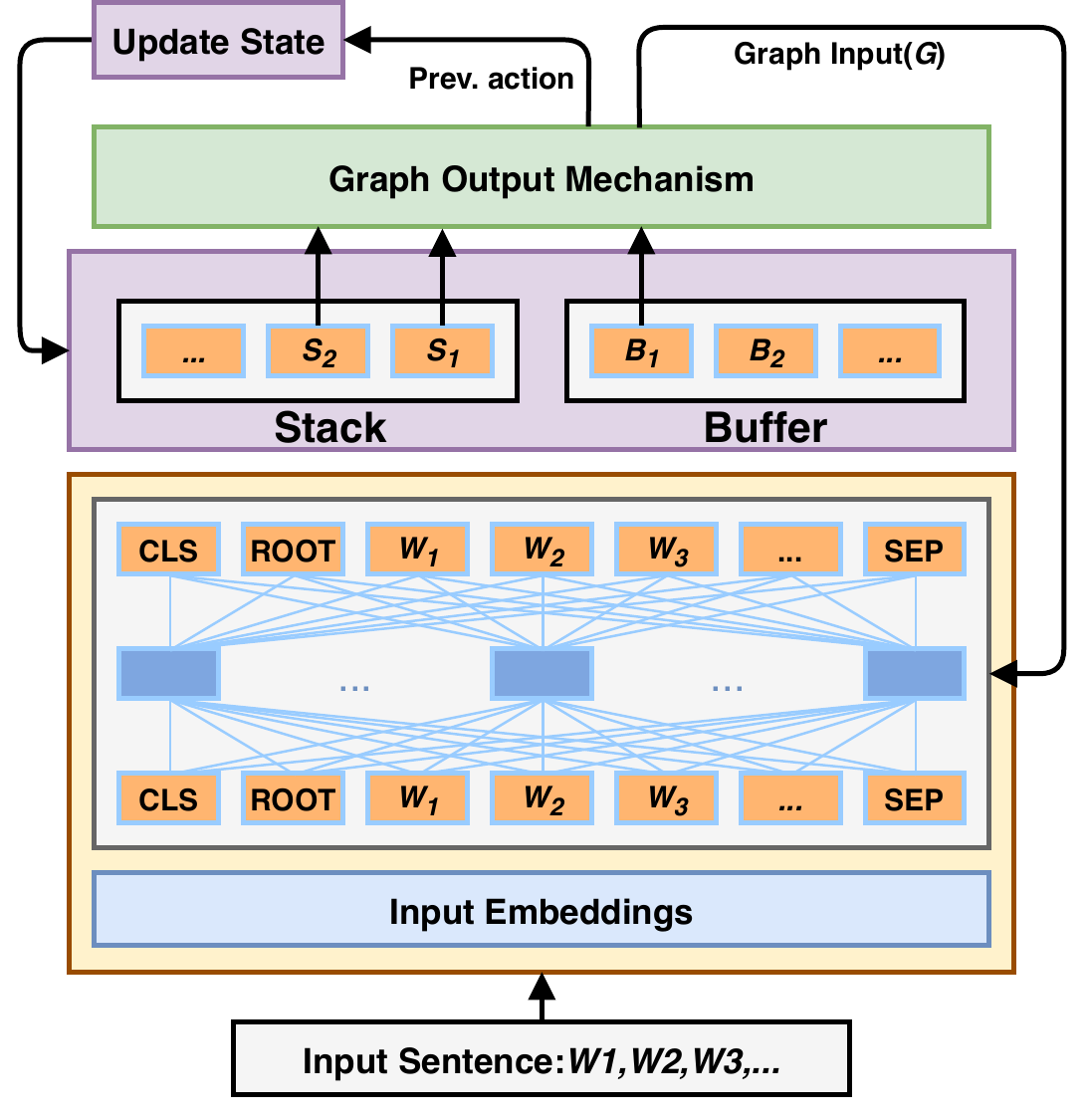}
                \vspace{-2ex}
		\caption{SentTr+G2GTr.\label{senttr}} 
	\end{subfigure}
        \vspace{-1ex}
	\caption{The State Transformer and Sentence Transformer parsers with Graph-to-Graph Transformer integrated.\label{fig:g2gtr}} 
\end{figure*}

We propose a version of the Transformer which is designed for both conditioning on graphs and predicting graphs, which we call Graph2Graph Transformer (G2GTr), and show how it can be applied to transition-based dependency parsing.  G2GTr supports arbitrary input graphs and arbitrary edges in the output graph.  But since the nodes of both these graphs are the input tokens, the nodes of the output graph are limited to the set of nodes in the input graph.

Inspired by the relative position embeddings of \newcite{shaw-etal-2018-self}, we use the attention mechanism of Transformer to input arbitrary graph relations.  By inputting the embedding for a relation label into the attention functions for the related tokens, the model can more easily learn to pass information between graph-local tokens, which gives the model an appropriate linguistic bias, without imposing hard constraints.

Given that the attention function is being used to input graph relations, it is natural to assume that graph relations can also be predicted with an attention-like function.  We do not go so far as to restrict the form of the prediction function, but we do restrict the vectors used to predict graph relations to only the tokens involved in the relation.

\subsection{Original Transformer}

Transformer~\cite{vaswani2017attention} is an encoder-decoder model, of which we only use the encoder component.  A Transformer encoder computes an output embedding for each token in the input sequence through stacked layers of multi-head self-attention.  
Each attention head takes its input vectors~($x_1,...,x_n$) and computes its output attention vectors~($z_1,...,z_n$).  Each $z_i \in R^m$ is a weighted sum of transformed input vectors $x_j \in R^m$:
\vspace{-0.5ex}
\begin{align}
\label{eq:tr1}
z_i = \sum_j\alpha_{ij}(x_j\boldsymbol{W^V})
\\[-5.5ex]\nonumber
\end{align}
with the attention weights
$\alpha_{ij} = \frac{\exp(e_{ij})}{\sum_{k=1}^n \exp(e_{ik})}$
and
\vspace{-0.5ex}
\begin{align}
\label{eq:tr3}
e_{ij} = \frac{(x_i\boldsymbol{W^Q})(x_j\boldsymbol{W^K})}{\sqrt{d}}
\\[-4.5ex]\nonumber
\end{align}
where $\boldsymbol{W^V}, \boldsymbol{W^Q}, \boldsymbol{W^K} \in R^{m \times d}$ are the trained value, query and key matrices, $m$ is the embedding size, and $d$ is the attention head size.

\subsection{Graph Inputs}
\label{graph-in}

Graph2Graph Transformer extends the architecture of the Transformer to accept any arbitrary graph as input.  In particular, we input the dependency tree as its set of dependency relations.  Each labelled relation $(x_i,x_j,l^\prime)$ is input by modifying Equation \ref{eq:tr3} as follows:
\vspace{-0.5ex}
\begin{align}
\label{eq:tr4}
e_{ij} = \frac{(x_i\boldsymbol{W^Q})(x_j\boldsymbol{W^K}+p_{ij}\boldsymbol{W^L_1})}{\sqrt{d}}
\\[-4.5ex]\nonumber
\end{align}
where $p_{ij}\in \{0,1\}^k$ is a one-hot vector which specifies the type $l^\prime$ of the relation between $x_i$ and $x_j$, discussed below, and $\boldsymbol{W^L_1} \in R^{k \times d}$ is a matrix of learned parameters. 
We also modify Equation \ref{eq:tr1} to transmit information about relations to the output of the attention layer:
\vspace{-0.5ex}
\begin{align}
\label{eq:tr5}
z_i = \sum_j\alpha_{ij}(x_j\boldsymbol{W^V}+p_{ij}\boldsymbol{W^L_2})
\\[-4.5ex]\nonumber
\end{align}
where $\boldsymbol{W^L_2} \in R^{k \times d}$ are learned parameters.

In this work, we consider graph input for only unlabelled directed dependency relations $l^\prime$, so $p_{ij}$ has only three dimensions ($k{=}3$), for \texttt{leftward}, \texttt{rightward} and \texttt{none}.  This choice was made mostly to simplify our extension of the Transformer, as well as to limit the computational cost of this extension.  The dependency labels are input as label embeddings added to the input token embeddings of the dependent word.

\subsection{Graph Outputs}
\label{graph-out}

The graph output mechanism of Graph2Graph Transformer predicts each labelled edge of the graph using the output embeddings of the tokens that are connected by that edge.  Because in this work we are investigating auto-regressive models, this prediction is done one edge at a time.  See \cite{Mohammadshahi_tacl20_forthcoming} for an investigation of non-autoregressive models using our G2GTr architecture.

In this work, the graph edges are labelled dependency relations, which are predicted as part of the actions of a transition-based dependency parser.  In particular, the Relation classifier uses the output embeddings of the top two elements on the stack and predicts the label of their dependency relation, conditioned on its direction.  There is also an Exist classifier, which uses the output embeddings of the top two elements on the stack and the front of the buffer to predict the type of parser action, {\tt SHIFT}, {\tt SWAP}, {\tt RIGHT-ARC}, or {\tt LEFT-ARC}.
\vspace{-0.5ex}
\begin{align}
\label{eq:classifier}
\begin{split}
&a^t = \operatorname{Exist}([g^t_{s_2},g^t_{s_1},g^t_{b_1}]) \\
&l^t = \operatorname{Relation}([g^t_{s_2},g^t_{s_1}]\:\: | \: a^t)
\end{split}
\\[-4.5ex]\nonumber
\end{align}
where $g^t_{s_2}$, $g^t_{s_1}$, and $g^t_{b_1}$ are the output embeddings of top two tokens in the stack and the front of buffer, respectively.  The $\operatorname{Exist}$ and $\operatorname{Relation}$ classifiers are
MLPs with one hidden layer.

For the transition-based dependency parsing task, the chosen parser action and dependency label are used both to update the current partial dependency structure and to update the parser state.

\section{Parsing Models}
In this section, we define two Transformer-based models for transition-based dependency parsing, and integrate the Graph2Graph Transformer architecture with them, as illustrated in Figure~\ref{fig:g2gtr}.

\subsection{State Transformer}

We propose a novel attention-based architecture, called State Transformer (StateTr), which computes a comprehensive representation for the parser state. Inspired by \newcite{dyer-etal-2015-transition}, we directly use the parser state, meaning both the stack and buffer elements, as the input to the Transformer model. We additionally incorporate components that have proved successful in \newcite{dyer-etal-2015-transition}. In the remaining paragraphs, we describe each component in more detail.

\subsubsection{Input Embeddings}

The Transformer architecture takes a sequence of input tokens and converts them into a sequence of input embedding vectors,  before computing its context-dependent token embeddings. For the State Transformer model, the sequence of input tokens represents the current parser state, as illustrated in Figure~\ref{statetr}.

\paragraph{Input Sequence:}
The input symbols include the words of the sentence $\Omega = (w_1,w_2,...,w_n)$ with their associated part-of-speech tags (PoS) $(\alpha_1,\alpha_2,...,\alpha_n)$.  Each of these words can appear in the stack or buffer of the parser state. Besides, there is the {\tt ROOT} symbol, for the root of the dependency tree, which is always on the bottom of the stack. Inspired by the input representation of BERT~\cite{bert:18}, we also use two special symbols, {\tt CLS} and {\tt SEP}, which indicate the different parts of the parser state.

The sequence of input tokens starts with the {\tt CLS} symbol, then includes the tokens on the stack from bottom to top.  Then it has a {\tt SEP} symbol, followed by the tokens on the buffer from front to back so that they are in the same order in which they appeared in the sentence. Given this input sequence, the model computes a sequence of vectors which are input to the Transformer network. Each vector is the sum of several embeddings, which are defined below.

\paragraph{Input Token Embeddings:}
The embedding of each token~($w_i$) is calculated as:
\vspace{-0.5ex}
\begin{align}
\label{eq:tokenembedding}
  T_{w_i} = \operatorname{Emb}(w_i) + \operatorname{Emb}(\alpha_i)
\\[-4ex]\nonumber
\end{align} 
where $\operatorname{Emb}(w_i), \operatorname{Emb}(\alpha_i) \in R^m$ are the word and PoS embeddings respectively.
For the word embeddings, we use the pre-trained word vectors of the BERT model. During training and evaluation, we use the pre-trained embedding of first sub-word as the token representation of each word and discard embeddings of non-first sub-words due to training efficiency.\footnote{Using embeddings of first sub-word for each word results in better performance than using the last one or averaging all of them as also shown in previous works~\cite{Kondratyuk_2019,kitaev-etal-2019-multilingual}.} The PoS embeddings are trained parameters. 

\paragraph{Composition Model:}

\begin{figure}
  \centering
  \includegraphics[width=0.5\linewidth]{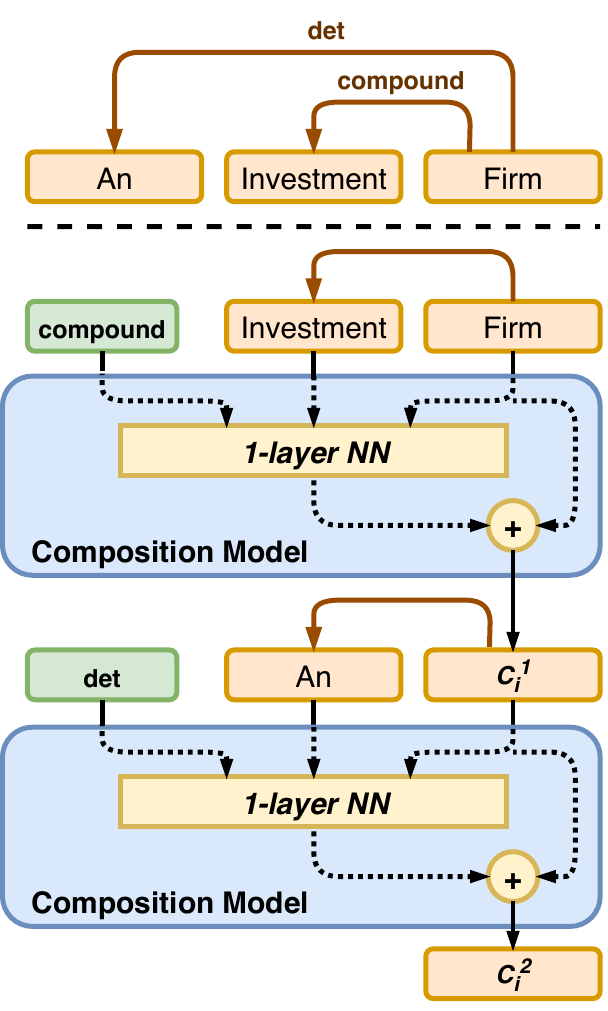}
  \vspace{-1ex}
  \caption{An Example of Composition model.
  }
  \label{fig:composition}
\end{figure}

As an alternative to our proposed graph input method, 
previous work has shown that complex phrases can be input to a neural network by using recursive neural networks to recursively compose the embeddings of sub-phrases~\cite{socher2011dynamic,socher2014grounded,socher-etal-2013-recursive,hermann-blunsom-2013-role,tai2015improved}. 
We extend the proposed composition model of \newcite{dyer-etal-2015-transition} by applying a one-layer feed-forward neural network as a composition model and adding skip connections to each recursive step.\footnote{These skip connections help address the vanishing gradient problem, and preliminary experiments indicated that they were necessary to integrate pre-trained BERT~\cite{bert:18} parameters with the model (discussed in Section~\ref{bert-pretrain} and Appendix~\ref{pre-exp-skip}).}
Since a syntactic head may contain an arbitrary number of dependents, we compute new token embeddings of head-dependent pairs one at a time as they are specified by the parser, as shown in Figure~\ref{fig:composition}.
At each parser step $t$, we compute each new token embedding $C_i^t$ of token $i$ by inputting to the composition model, its previous token embedding $C_j^{t-1}$ and the embedding of the most recent dependent with its associated dependency label, where $j$ is the position of token $i$ in the previous parser state.
At $t=0$, $C_i^0$ is set to the initial token embedding $T_{w_i}$.
More mathematical and implementation details are given in Appendix~\ref{sup:compmodel}.

\paragraph{Position and Segment Embeddings:}

To distinguish the different positions and roles of words in the parser state, we add their embeddings to the token embeddings.  Position embeddings $\beta_i$ encode the token's position in the whole sequence.\footnote{Preliminary experiments showed that using position embeddings for the whole sequence achieves better performance than applying separate position embeddings for each segment (More detail in Appendix~\ref{pre-exp-pos}).}  Segment embeddings $\gamma_i$ encode that the input sequence contains distinct segments (e.g.\ stack and buffer).  

\paragraph{Total Input Embeddings:}
Finally, at each step $t$, we sum the outputs of the composition model with the segment and position embeddings and consider them as the sequence of input embeddings for our State Transformer model.
\vspace{-0.5ex}
\begin{align}
\label{eq:finalinput}
x^t_i = C^t_i + \gamma_i + \beta_i 
\end{align}

\subsubsection{History Model}

We define a history model similar to \newcite{dyer-etal-2015-transition}, to capture the information about previously specified transitions. The output $h^t$ of the history model is computed as follows:
\vspace{-0.5ex}
\begin{align}
\label{eq:history}
h^t,~c^t = \operatorname{LSTM}((h^{t-1},c^{t-1}),a^t+l^t) 
\\[-4.5ex]\nonumber
\end{align}
where $a^t$ and $l^t$ are the previous transition and its associated dependency label, and $h^{t-1}$ and $c^{t-1}$ are the previous output vector and cell state of the history model.
The output of the history model is input directly to the parser action classifiers in (\ref{eq:classifier}).

\subsection{Sentence Transformer}

We propose another attention-based architecture, called Sentence Transformer (SentTr), to compute a representation for the parser state.  This model first uses a Transformer to compute context-dependent embeddings for the tokens in the input sentence.  Similarly to \newcite{cross-huang-2016-incremental}, a separate stack and buffer data structure is used to keep track of the parser state, as shown in Figure~\ref{senttr}, and the context-dependent embeddings of the tokens that are involved in the next parser action are used to predict the next transition.
More specifically, the input sentence tokens are computed with the BERT tokeniser~\cite{bert:18} and the next transition is predicted from the embeddings of the first sub-words of the top two elements of the stack and the front element of the buffer.\footnote{Predicting transitions with the embedding of first sub-word for each word results in better performance than using the last one or all of them as also shown in previous works.~\cite{Kondratyuk_2019,kitaev-etal-2019-multilingual}}

In the baseline version of this model, the Transformer which computes the token embeddings does not see the structure of the parser state nor the partial dependency structure.

In Sentence Transformer, the sequence of input tokens starts with a {\tt CLS} token and ends with a {\tt SEP} token, as in the BERT~\cite{bert:18} input representation. It also includes the {\tt ROOT} symbol for the root of the dependency tree. The input embeddings are derived from input tokens as:
\vspace{-0.5ex}
\begin{align}
\label{eq:senttr-input}
x_{i} = \operatorname{Emb}(w_i)+\operatorname{Emb}(\alpha_i)+\beta_i
\\[-4.5ex]\nonumber
\end{align}
where $x_i$ is the input embedding for token $w_i$, $\operatorname{Emb}(.)$ is defined as in Equation~(\ref{eq:tokenembedding}), and $\beta_i$ is the positional embedding for the element at position $i$.

\subsection{Integrating with G2G Transformer}

We use the two proposed attention-based dependency parsers above as baselines, and evaluate the effects of integrating them with the Graph2Graph Transformer architecture.  We modify the encoder component of each baseline model by adding the graph input mechanism defined in Section~\ref{graph-in}. Then, we compute the new partially constructed graph as follows:
\vspace{-0.5ex}
\begin{align}
\label{eq:integ-g2g}
\begin{split}
&Z^t = \operatorname{Gin}(X,G^{t}) \\
&G^{t+1} = G^{t}\cup\operatorname{Gout}(\operatorname{Select}(Z^t,P^t))
\end{split}
\\[-4.5ex]\nonumber
\end{align}
where $G^{t}$ is the current partially specified graph, $Z^t$ is the encoder's sequence of output token embeddings, $P^t$ is the parser state, and $G^{t+1}$ is the newly predicted partial graph. $\operatorname{Gin}$, and $\operatorname{Gout}$ are the graph input and graph output mechanisms defined in Sections~\ref{graph-in} and \ref{graph-out}. The $\operatorname{Select}$ function selects from $Z^t$, the token embeddings of the top two elements on the stack and the front of the buffer, based on the parser state $P^t$. 
More specifics about each baseline are given in the following paragraphs.\footnote{A worked example of both baseline models integrated with G2GTr is provided in Appendix~\ref{ap:example}.}

\paragraph{State Tr +G2GTr:}
To input all the dependency relations in the current partial parse, we add a third segment to the parser state, called the Deleted list $D$, which includes words that have been removed from the buffer and stack after having both their children and parent specified. The order of words in $D$ is the same as the input sentence.  The current partial dependency structure is then input with the graph input mechanism as relations between the words in this extended parser state.
To show the effectiveness of the graph input mechanism, we exclude the composition model from the State Transformer model when integrated with the Graph2Graph Transformer architecture. 
We will demonstrate the impact of this replacement in Section~\ref{sec:penn-result}.

\paragraph{Sentence Tr +G2GTr:}
The current partial dependency structure is input with the graph input mechanism as relations between the first sub-words of the head and dependent words of each dependency relation.
For the non-first subwords of each word, we define a new dependency relation with these subwords dependent on their associated first sub-word.

\subsection{Pre-Training with BERT}
\label{bert-pretrain}

Initialising a Transformer model with the pre-trained parameters of BERT~\cite{bert:18}, and then fine-tuning on the target task, has demonstrated large improvements in many tasks.  But our version of the Transformer has novel inputs that were not present when BERT was trained, namely the graph inputs to the attention mechanism and the composition embeddings (for State Transformer).  Also, the input sequence of State Transformer has a novel structure, which is only partially similar to the input sentences which BERT was trained on.  So it is not clear that BERT pre-training will even work with this novel architecture.
To evaluate whether BERT pre-training works for our proposed architectures,
we also initialise the weights of our models with the first $n$ layers of BERT, where $n$ is the number of self-attention layers in the model.

\section{Experimental Setup}

\subsection{Datasets}

We evaluate our models on two types of datasets, WSJ Penn Treebank, and Universal Dependency (UD) Treebanks. Following \newcite{kulmizev2019deep}, for evaluation, we include punctuation for UD treebanks and exclude it for the WSJ Penn Treebank~\cite{nilsson-nivre-2008-malteval}.\footnote{Description of Treebanks are provided in Appendix~\ref{sup:treebank}.}

\paragraph{WSJ Penn Treebank:}
We train our models on the Stanford dependency version of the English Penn Treebank~\cite{marcus-etal-1993-building}. We use the same setting as defined in \newcite{dyer-etal-2015-transition}. We additionally add section 24 to our development set to avoid over-fitting. For PoS tags, we use Stanford PoS tagger~\cite{toutanova-etal-2003-feature}. 

\paragraph{Universal Dependency Treebanks:}
We also train models on Universal Dependency Treebanks (UD v2.3)~\cite{11234/1-2895}. We evaluate our models on the list of languages defined in \newcite{kulmizev2019deep}. This set of languages contains different scripts, various morphological complexity and character
set sizes, different training sizes, and non-projectivity ratios.

\subsection{Models}

As strong baselines from previous work, we compare our models to previous traditional transition-based and Seq2Seq models.  For a fair comparison with previous models, we consider ``traditional'' transition-based parsers to be those that predict a fixed set of scores for each decoding step.\footnote{We do not consider the models of \cite{ma-etal-2018-stack,FerGomNAACL2019} to be comparable to traditional transition-based models like ours because they make decoding decisions between $O(n)$ alternatives.  In this sense, they are in between the $O(1)$ alternatives for transition-based models and the $O(n^2)$ alternatives for graph-based models.  Future work will investigate applying Graph2Graph Transformer to these types of parsers as well.}

To investigate the usefulness of each component of the proposed parsing models, we evaluate several versions. For the State Transformer, we evaluate StateTr and StateTr+G2GTr models both with and without BERT initialisation. To further analyse the impact of Graph2Graph Transformer, we also compare to keeping the composition function of the StateTr model when integrated with G2GTr~(StateTr+G2GTr+C). To further demonstrate the impact of the graph output mechanism, we compare to using the output embedding of the {\tt CLS} token as the input to the transition classifiers for both the baseline model (StateCLSTr) and its combined version~(StateTr+G2CLSTr). For Sentence Transformer, we evaluate the SentTr and SentTr+G2GTr models with BERT initialisation. 
We also evaluate the best variations of each baseline on the UD Treebanks.\footnote{The number of parameters and average running times for each model are provided in Appendix~\ref{sup:run-detail}.}

\subsection{Details of Implementation}

All hyper-parameter details are given in Appendix~\ref{sup:hyper-par}. Unless specified otherwise, all models have 6 self-attention layers. We use the AdamW optimiser provided by \newcite{Wolf2019HuggingFacesTS} to fine-tune model parameters.
All our models use greedy decoding, meaning that at each step only the highest scoring parser action is considered for continuation.  This was done for simplicity, although beam search could also be used. The pseudo-code for computing the elements of the graph input matrix ($p_{ij}$) for each baseline is provided in Appendix~\ref{sup:graph-input}.

\section{Results and Discussion}
\label{sec:penn-result}

\subsection{English Penn Treebank Result}

\begin{table}[t]
    \footnotesize
	\centering
	\tabcolsep=0.1cm
	\begin{adjustbox}{width=\linewidth}
	\begin{tabular}{@{}l@{~}ll@{~}c@{~}ll@{}}\toprule
		 & \multicolumn{2}{c}{Dev Set} & \phantom{a}
	         & \multicolumn{2}{c}{Test Set} 
                 \\
		 \cmidrule{2-3} \cmidrule{5-6}
		 & ~UAS & ~LAS & & ~UAS & ~LAS \\ \midrule
        \hline 
		\textbf{Transition-based:}\\
		\newcite{dyer-etal-2015-transition} & & && {93.10} & {90.90} \\
		\newcite{weiss-etal-2015-structured} & & && 94.26 & 91.42 \\
		\newcite{cross-huang-2016-incremental} & & && 93.42 & 91.36 \\
        \newcite{ballesteros2016training} & & && 93.56 & 92.41 \\
		\newcite{andor-etal-2016-globally} & & && 94.61 & 92.79 \\
		\newcite{kiperwasser-goldberg-2016-simple}\hspace{-5ex} & & && 93.90 & 91.90 \\
		\newcite{yang2017joint} & & && 94.18 & 92.26 \\
		\hline
		\textbf{Seq2Seq-based:} \\
		\newcite{zhang-etal-2017-stack} & & && 93.71 & 91.60 \\
		\newcite{li-etal-2018-seq2seq} & & && 94.11 & 92.08 \\
		\hline
		StateTr & 91.94 & 89.07 && 92.32 & 89.69 \\
		StateTr+G2GTr & 92.53 & 90.16 && 93.07 & 91.08 \\[0.5ex]
		BERT StateTr & 94.66 & 91.94 && 95.18 & 92.73 \\
		BERT StateCLSTr & 93.62 & 90.95 && 94.31 & 91.85 \\
		BERT StateTr+G2GTr & \textbf{94.96} & \textbf{92.88} && \textbf{95.58} & \textbf{93.74} \\
		BERT StateTr+G2CLSTr & 94.29 & 92.13 && 94.83 & 92.96 \\
		BERT StateTr+G2GTr+C & 94.41 & 92.25 && 94.89 & 92.93 \\
		\hline
		BERT SentTr & 95.34 & 93.29 && 95.65 & 93.85 \\
		BERT SentTr+G2GTr & \textbf{95.66} & \textbf{93.60} && \textbf{96.06} & \textbf{94.26} \\
		BERT SentTr+G2GTr-7 layer & \textbf{95.78} & \textbf{93.74} && \textbf{96.11} & \textbf{94.33} \\
		\bottomrule
	    \end{tabular}
	\end{adjustbox}
        \vspace{-1.5ex}
	\caption{\label{tab:wsj-prev} 
	  Comparisons to SoTA on English WSJ Treebank Stanford dependencies.
        }
\end{table}

\begin{table}[t]
	\centering
	
	\begin{adjustbox}{width=\linewidth}
	\begin{tabular}{|@{~~}l@{~}|@{~~}c@{~}|@{~~}c@{~}|@{~~}c@{~}|}
		\hline
		\!Language & \parbox{\widthof{Kulmizev et al.}}{\addtolength{\baselineskip}{-0.5ex}~\\ \newcite{kulmizev2019deep}} & \parbox{\widthof{StateTr+G2GTr}}{\addtolength{\baselineskip}{-0.5ex}~\\ BERT StateTr+G2GTr} & \parbox{\widthof{SentTr+G2GTr}}{\addtolength{\baselineskip}{-0.5ex}~\\ BERT SentTr+G2GTr}\\[1.5ex]
        \hline
         \vspace{-2ex}
         & & & \\
		Arabic & {81.9} & {82.63} & {\textbf{83.65}} \\
		Basque & {77.9} & {74.03} & {\textbf{83.88}} \\ 
		Chinese & {83.7} & {85.91} & {\textbf{87.49}} \\
		English & {87.8} & {89.21} & {\textbf{90.35}} \\
		Finnish & {85.1} & {80.87} & {\textbf{89.47}} \\
		Hebrew & {85.5} & {87.0} & {\textbf{88.75}} \\
		Hindi & {89.5} & {93.13} & {\textbf{93.12}} \\
		Italian & {92.0} & {92.6} & {\textbf{93.99}} \\
		Japanese & {92.9} & {95.25} & {\textbf{95.51}} \\
		Korean & {83.7} & {80.13} & {\textbf{87.09}} \\
		Russian & {91.5} & {92.34} & {\textbf{93.30}} \\
		Swedish & {87.6} & {88.36} & {\textbf{90.40}} \\
		Turkish & {64.2} & {56.87} & {\textbf{67.77}} \\
		\hline
		Average & {84.87} & {84.48} & {\textbf{88.06}} \\
		\hline
	\end{tabular}
	\end{adjustbox}
	\caption{\label{tab:ud-best} Labelled attachment score on 13 UD corpora for \newcite{kulmizev2019deep} with BERT pre-training, BERT StateTr+G2GTr, and BERT SentTr+G2GTr models.
          \vspace{-1ex}
        } 
\end{table}

In Table~\ref{tab:wsj-prev}, we show several variations of our models, and previous state-of-the-art transition-based and Seq2Seq parsers on WSJ Penn Treebank.\footnote{Results are calculated with the official evaluation script provided in \url{https://depparse.uvt.nl/}.} 
For State Transformer, replacing the composition model (StateTr) with our graph input mechanism (StateTr+G2GTr) results in 9.97\%\,/\,11.66\% LAS relative error reduction (RER) without\,/\,with BERT initialisation,
which demonstrates its effectiveness. Comparing to the closest previous model for conditioning of the parse graph, the StateTr+G2GTr model reaches better results than the StackLSTM model~\cite{dyer-etal-2015-transition}. Initialising our models with pre-trained BERT achieves 26.25\% LAS RER for the StateTr model, and 27.64\% LAS RER for the StateTr+G2GTr model, thus confirming the compatibility of our G2GTr architecture with pre-trained Transformer models. The BERT StateTr+G2GTr model outperforms previous state-of-the-art models.
Removing the graph output mechanism (StateCLSTr\,/\,StateTr+G2CLSTr) results in a 12.28\%\,/\,10.53\% relative performance drop for the StateTr and StateTr+G2GTr models, respectively, which demonstrates the importance of our graph output mechanism.  
If we consider both the graph input and output mechanisms together, adding them both (BERT StateTr+G2GTr) to BERT StateCLSTr achieves 21.33\% LAS relative error reduction, which shows the synergy of using both mechanisms together.
But then adding the composition model (BERT StateTr+G2GTr+C) results in an 8.84\% relative drop in performance, which demonstrates again that our proposed graph input method is a more effective way to model the partial parse than recursive composition models. 

For Sentence Transformer, the synergy between its encoder and BERT results in excellent performance even for the baseline model (compared to \newcite{cross-huang-2016-incremental}).  Nonetheless, adding G2GTr achieves significant improvement (4.62\% LAS RER), which again demonstrates the effectiveness of the Graph2Graph Transformer architecture.  
Finally, we also evaluate the BERT SentTr+G2GTr model with 7 self-attention layers instead of 6, resulting in 2.19\% LAS RER, which motivates future work on larger Graph2Graph Transformer models.

\subsection{UD Treebanks Results}

In Table~\ref{tab:ud-best}, we show LAS scores on 13 UD Treebanks\footnote{Unlabelled attachment scores, and results of development set are provided in the Appendix~\ref{suptab:scoreud-uas}. Results are calculated with the official UD evaluation script (\url{https://universaldependencies.org/conll18/evaluation.html}).}.
As the baseline, we use scores of the transition-based model proposed by \newcite{kulmizev2019deep}, which uses the deep contextualized word representations of BERT and ELMo~\cite{Peters_2018} as an additional input to their parsing models. Our BERT~StateTr+G2GTr model outperforms the baseline on 9 languages, again showing the power of the G2GTr architecture.
But for morphology-rich languages such as Turkish and Finish, the StateTr parser design choice of only inputting the first sub-word of each word causes too much loss of information, resulting in lower results for our BERT~StateTr+G2GTr model than the baseline.
This problem is resolved by our SentTr parser design because all sub-words are input.  The BERT~SentTr+G2GTr model performs substantially better than the baseline on all languages, which confirms the effectiveness of our Graph2Graph Transformer architecture to capture a diversity of types of structure from a variety of corpus sizes.

\subsection{Error Analysis}

\begin{figure*}
  \centering
  \hspace{-2ex}
  \includegraphics[width=0.33\linewidth]{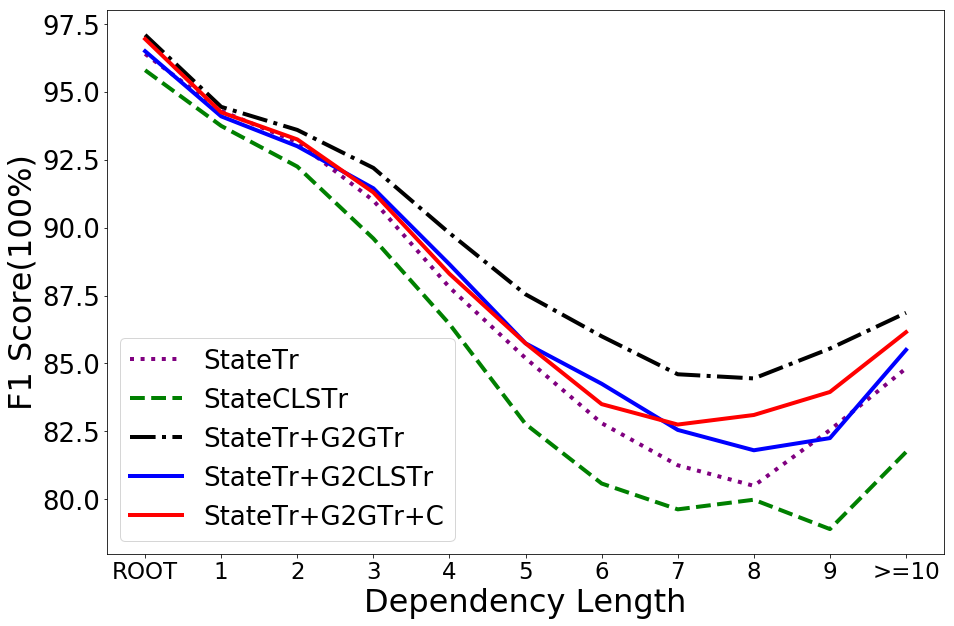}
  \includegraphics[width=0.33\linewidth]{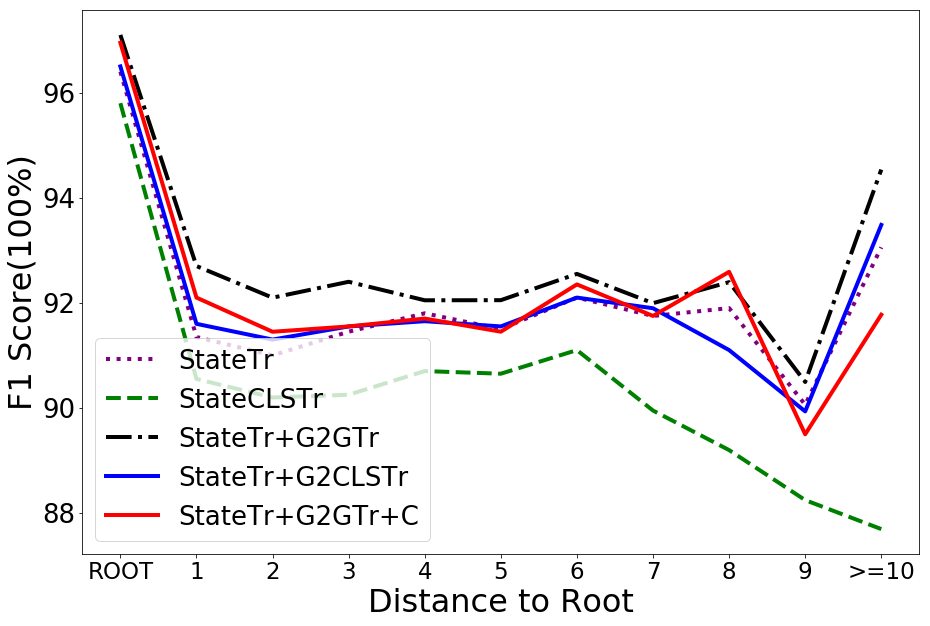}
  \includegraphics[width=0.33\linewidth]{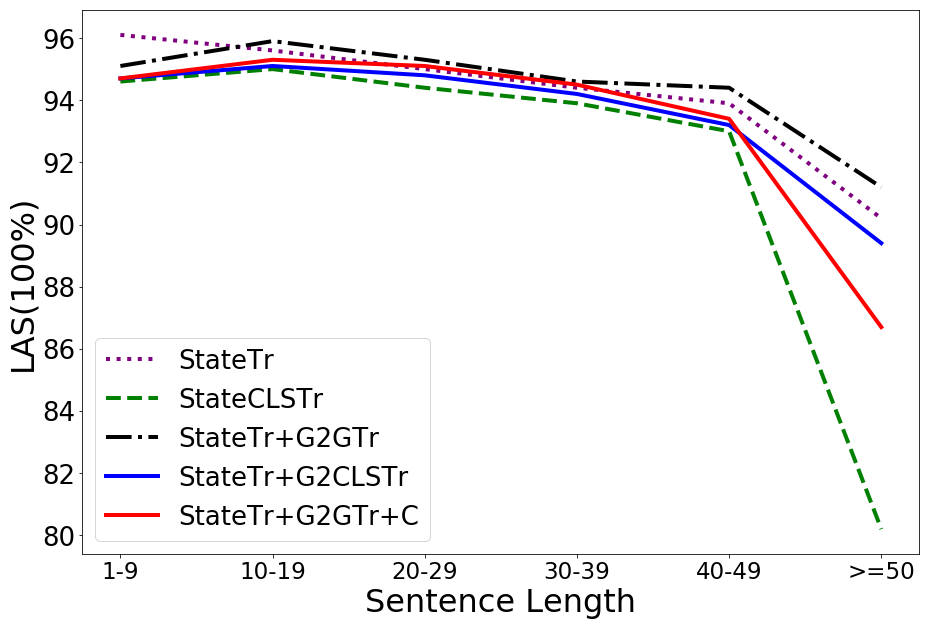}
  \hspace{-2ex}
  \caption{Error analysis of our models on the development set of the WSJ dataset.
    \vspace{-1ex}
  }
  \label{fig:errors}
\end{figure*}

To analyse the effectiveness of the proposed graph input and output mechanisms in variations of our StateTr model pre-trained with BERT, we follow \newcite{mcdonald-nivre-2011-analyzing} and measure their accuracy as a function of dependency length, distance to root, sentence length, and dependency type, as shown in Figure~\ref{fig:errors} and Table~\ref{table:acc-deptype}.\footnote{We use MaltEval\cite{nilsson-nivre-2008-malteval} tool for computing accuracies. Tables of results for the error analysis in Figure~\ref{fig:errors}, and Table~\ref{table:acc-deptype} are in the Appendix~\ref{appendix:error}.}. These results demonstrate that most of the improvement of the StateTr+G2GTr model over other variations comes from the hard cases which require a more global view of the sentence. 

\paragraph{Dependency Length:} The leftmost plot shows labelled F-scores on dependencies binned by dependency lengths. The integrated G2GTr models outperform other models on the longer (more difficult) dependencies, which demonstrates the benefit of adding the partial dependency tree to the self-attention model, which provides a global view of the sentence when the model considers long dependencies. Excluding the graph output mechanism also results in a drop in performance particularly in long dependencies. Keeping the composition component in the StateTr+G2GTr model doesn't improve performance at any length.

\paragraph{Distance to Root:} The middle plot shows the labelled F-score for dependencies binned by the distance to the root, computed as the number of dependencies in the path from the dependent to the root node. The StateTr+G2GTr models outperform baseline models on nodes that are of middle depths, which tend to be neither near the root nor near the leaves, and thus require more global information, as well as deeper nodes.

\paragraph{Sentence Length:} The rightmost plot shows labelled attachment scores (LAS) for sentences with different lengths. The relative stability of the StateTr+G2GTr model across different sentence lengths again shows the effectiveness of the Graph2Graph Transformer model on the harder cases.  Not using the graph output method shows particularly bad performance on long sentences, as does keeping the composition model.

\begin{table}
\centering
  \begin{adjustbox}{width=\linewidth}
    \begin{tabular}{|@{~}c@{~}|l@{~}|l|l|}
    \hline
    {\small Type} & {\small StateTr+G2GTr} & {\small StateTr} & {\small StateTr+G2CLSTr} \\
    \hline
\rule{0pt}{2.5ex} 
{\tt rcmod} & 86.84 & 76.38 (-79.5\%)  & 83.91  (-22.3\%) \\ \hline
\rule{0pt}{2.5ex} 
{\tt nsubjpass} & 95.49 & 92.70 (-61.9\%)  & 94.08  (-31.1\%)  \\ \hline
\rule{0pt}{2.5ex} 
{\tt ccomp} & 89.49 & 81.82 (-73.0\%)  & 87.56  (-18.4\%)  \\ \hline
\rule{0pt}{2.5ex} 
{\tt infmod} & 87.38 & 79.19 (-64.9\%)  & 84.93  (-19.4\%)  \\ \hline
\rule{0pt}{2.5ex} 
{\tt neg} & 95.75 & 94.84 (-21.4\%)  & 93.78  (-46.2\%)  \\ \hline
\rule{0pt}{2.5ex} 
{\tt csubj} & 76.94 & 67.93 (-39.0\%)  & 70.83  (-26.5\%)  \\ \hline
\rule{0pt}{2.5ex} 
{\tt cop} & 93.08 & 92.62 (-6.5\%)  & 91.58  (-21.7\%)  \\ \hline
\rule{0pt}{2.5ex} 
{\tt cc} & 90.90 & 90.45 (-4.9\%)  & 88.80  (-23.1\%)  \\
    \hline
    \end{tabular}
  \end{adjustbox}
\caption{F-scores (and RER) of our full BERT model (StateTr+G2GTr), without graph inputs (StateTr), and without graph outputs (StateTr+G2CLSTr) for some dependency types on the development set of WSJ Treebank, ranked by total negative RER.  Relative error reduction is computed w.r.t.\ the StateTr+G2GTr scores.}
\label{table:acc-deptype}
\end{table}

\paragraph{Dependency Type:} 
Table~\ref{table:acc-deptype} shows F-scores of different dependency types.
Excluding the graph input (StateTr) or graph output (StateTr+G2CLSTr) mechanisms results in a substantial drop for many dependency types, especially hard cases where accuracies are relatively low, and cases such as {\tt ccomp} which require a more global view of the sentence. 

\section{Conclusion}

We proposed the Graph2Graph Transformer architecture, which inputs and outputs arbitrary graphs through its attention mechanisms.  Each graph relation is input as a label embedding to each attention function involving the relation's tokens, and each graph relation is predicted from its token's embeddings like an attention function.  
We demonstrate the effectiveness of this architecture on transition-based dependency parsing,
where the input graph is the partial dependency structure specified by the parse history, and the output graph is predicted one dependency at a time by the parser actions.

To establish strong baselines, we also propose two Transformer-based models for this task, called State Transformer and Sentence Transformer. The former model incorporates history and composition models, as proposed in previous work.
Despite the competitive performance of these extended-Transformer parsers, adding our graph input and output mechanisms results in significant improvement.
Also, the graph inputs are effective replacements for the composition models.
All these results are preserved with the incorporation of BERT pre-training, which results in substantially improving the state-of-the-art in traditional transition-based dependency parsing.

As well as the generality of the graph input mechanism, the generality of the graph output mechanism means that Graph2Graph Transformer can be integrated with a wide variety of decoding algorithms.  For example, \newcite{Mohammadshahi_tacl20_forthcoming} investigate non-autoregressive decoding, which addresses the computational cost of running the G2GTr model once for every dependency edge.  Graph2Graph Transformer can also easily be applied to a wide variety of NLP tasks, such as semantic parsing tasks, which we hope to demonstrate in future work. 
\section*{Acknowledgement}

We are grateful to the Swiss NSF, grant CRSII5\_180320,
for funding this work.
We also thank Lesly Miculicich, other members of the IDIAP NLU group, and anonymous reviewers for helpful discussions.


\newpage

\bibliographystyle{acl_natbib}
\bibliography{anthology,emnlp2020}

\newpage
\renewcommand\thesection{\Alph{section}}
\renewcommand\thesubsection{\thesection.\Alph{subsection}}
\setcounter{section}{0}
\setcounter{table}{0}
\onecolumn
\begin{appendices}

\section{Preliminary Experiments}

\subsection{Skip Connection}
\label{pre-exp-skip}
Skip connections of composition model help address the vanishing gradient problem, and following experiments show that they are necessary to integrate pre-trained BERT~\cite{bert:18} parameters with the model:
\begin{table}[!ht]
    \footnotesize
	\centering
	
	\begin{adjustbox}{width=0.4\linewidth}
	\begin{tabular}{|l|c|c|}
		\hline
		Model & UAS & LAS \\
        \hline
		BERT StateTr  & 94.78 & 92.06 \\
		BERT StateTr without skip & 93.16 & 90.51 \\ 
		\hline
	    \end{tabular}
	\end{adjustbox}
	\caption{Preliminary experiments on the development set of WSJ Penn Treebank for BERT StateTr model with/without skip connections.
        } 
\end{table}

\subsection{Position Embeddings}
\label{pre-exp-pos}

Following experiments show that using position embeddings for the whole sequence achieves better performance than applying separate position embeddings for each segment:
\begin{table}[!ht]
    \footnotesize
	\centering
	
	\begin{adjustbox}{width=0.45\linewidth}
	\begin{tabular}{|l|c|c|}
		\hline
		Model & UAS & LAS \\
        \hline
		BERT StateTr  & 94.78 & 92.06 \\
		BERT StateTr with separate pos & 93.10 & 90.39 \\ 
		\hline
	    \end{tabular}
	\end{adjustbox}
	\caption{Preliminary experiments on the development set of WSJ Penn Treebank for BERT StateTr model, and its variation with separate position embeddings for each section.
        } 
\end{table}

\section{Composition Model}
\label{sup:compmodel}

Previous work has shown that recursive neural networks are capable of inducing a representation for complex phrases by recursively embedding sub-phrases~\cite{socher2011dynamic,socher2014grounded,socher-etal-2013-recursive,hermann-blunsom-2013-role}.
\newcite{dyer-etal-2015-transition} showed that this is an effective technique for embedding the partial parse subtrees specified by the parse history in transition-based dependency parsing.
Since a word in a dependency tree can have a variable number of dependents, they combined the dependency relations incrementally as they are specified by the parser.

We extend this idea by using a feed-forward neural network with {\tt Tanh} as the activation function and skip connections.
For every token in position $i$ on the stack or buffer, after deciding on step $t$, the composition model computes a vector $C^{t{+}1}_{a,i}$ which is added to the input embedding for that token:
\begin{align}
\label{eq:composition}
\begin{split}
&C^{t+1}_{a,i} = \operatorname{Comp}((\psi^t_{a,i},\omega^t_{a,i},{l^t_{a,i}})) + \psi^{t}_{a,i} \\
&\text{where:} a \in \{S,B\}
\end{split}
\end{align}
where the $\operatorname{Comp}$ function is a one-layer feed forward neural network, and $(\psi^t_{a,i},\omega^t_{a,i},{l^t_{a,i}})$ represents the most recent dependency relation  with head $\psi^t_{a,i}$ specified by the decision at step $t$ for element in position $i$ in the stack or buffer. 
In arc-standard parsing, the only word which might have received a new dependent by the previous decision is the word on the top of the stack, $i{=}1$.  This gives us the following definition of $(\psi^t_{a,i},\omega^t_{a,i},{l^t_{a,i}})$:
\begin{equation}
    \begin{cases}
      {\tt RIGHT{-}ARC}(l):\\
      \begin{array}{l}
        \psi^t_{S,1}{=}C^{t}_{S,2},~ \omega^t_{S,1}{=}C^{t}_{S,1},~ l^t_{S,1} {=} l,\\
        
        \psi^t_{S,i{\neq}1} {=} C^t_{S,i+1},~ \omega^t_{S,i{\neq}1} {=} \omega^t_{S,i+1},~ l^t_{S,i{\neq}1} {=} l^t_{S,i+1}\\

        \psi^t_{B,i} {=} C^t_{B,i} \\
        
        \end{array}\\
      {\tt LEFT{-}ARC}(l):\\
      \begin{array}{l}
        \psi^t_{S,1}{=}C^{t}_{S,1},~ \omega^t_{S,1}{=}C^{t}_{S,2},~ l^t_{S,1} {=} l,\\
        
        \psi^t_{S,i{\neq}1} {=} C^t_{S,i+1},~ \omega^t_{S,i{\neq}1} {=} \omega^t_{S,i+1},~ l^t_{S,i{\neq}1} {=} l^t_{S,i+1} \\
        
        \psi^t_{B,i} {=} C^t_{B,i} \\
        
      \end{array}\\
      
      {\tt SHIFT}:\\
      \begin{array}{l}
        \psi^t_{S,1} {=} C^t_{B,1},~ \omega^t_{S,1} {=} \omega^t_{B,1},~ l^t_{S,1} {=} l^t_{B,1} \\
        
        \psi^t_{S,i{\neq}1} {=} C^t_{S,i-1},~ \omega^t_{S,i{\neq}1} {=} \omega^{t}_{S,i-1},~ l^t_{S,i{\neq}1} {=} l^{t}_{S,i-1} \\
        
        \psi^t_{B,i} {=} C^t_{B,i+1},~ \omega^t_{B,i} {=} \omega^{t}_{B,i+1},~ l^t_{B,i} {=} l^{t}_{B,i+1} \\
      \end{array} \\

      {\tt SWAP}:\\
      \begin{array}{l}
      
        \psi^t_{S,1}{=}C^{t}_{S,1} \\ 
     
        \psi^t_{S,i{\neq}1} {=} C^t_{S,i+1},~ \omega^t_{S,i{\neq}1} {=} \omega^t_{S,i+1},~ l^t_{S,i{\neq}1} {=} l^t_{S,i+1} \\
        
        \psi^t_{B,1} {=} C^t_{S,2},~ \omega^t_{B,1} {=} \omega^t_{S,2},~ l^t_{B,1} {=} l^t_{S,2} \\
        
        \psi^t_{B,i{\neq}1} {=} C^t_{B,i-1},~ \omega^t_{B,i{\neq}1} {=} \omega^{t}_{B,i-1},~ l^t_{B,i{\neq}1} {=} l^{t}_{B,i-1} \\
      \end{array}
      
    \end{cases}
\end{equation}
where $C^{t}_{S,1}$ and $C^{t}_{S,2}$ are the embeddings of the top two elements of the stack at time step $t$, and $C^t_{B,1}$ is the embedding of the word on the front of the buffer at time $t$.  ${l^t_{a,i}}\in R^m$ is the label embedding of the specified relation, including its direction.
For all words which have not received a new dependent, the composition is computed anyway with the most recent dependent and label (with a {\tt [NULL]} dependent and label of that position{\tt [L{-}NULL]} if there is no dependency relation with element $i$ as the head).\footnote{Preliminary experiments indicated that not updating the composition embedding for these cases resulted in worse performance.}

At $t=0$, $C^t_{a,i}$ is set to the initial token embedding $T_{w_i}$.
The model then computes Equation~\ref{eq:composition} iteratively at each step $t$ for each token on the stack or buffer.

There is a skip connection in Equation~\ref{eq:composition} to address the vanishing gradient problem.
Also, preliminary experiments showed that without this skip connection to bias the composition model towards the initial token embeddings $T_{w_i}$, integrating pre-trained BERT~\cite{bert:18} parameters into the model did not work.
\vspace{-1ex}

\newpage
\section{Example of the Graph-to-Graph Transformer parsing}
\label{ap:example}

\begin{figure*}[!ht]
\begin{subfigure}{.5\textwidth}
  \centering
  \includegraphics[width=0.9\textwidth,height=0.75\textheight]{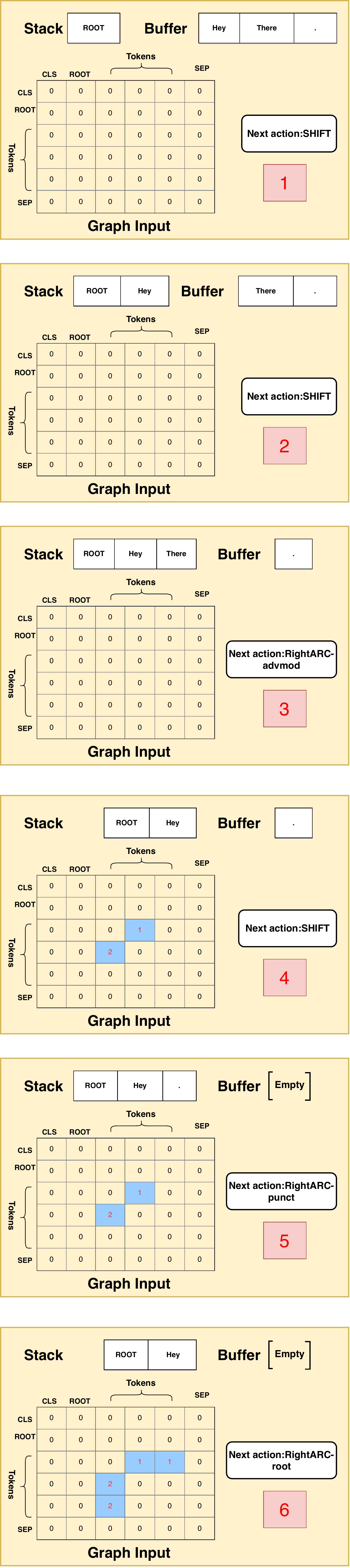}  
  \caption{SentTr+G2G-Tr.}
  \label{fig:senttr}
\end{subfigure}
\begin{subfigure}{.5\textwidth}
  \centering
  \includegraphics[width=0.92\textwidth,height=0.75\textheight]{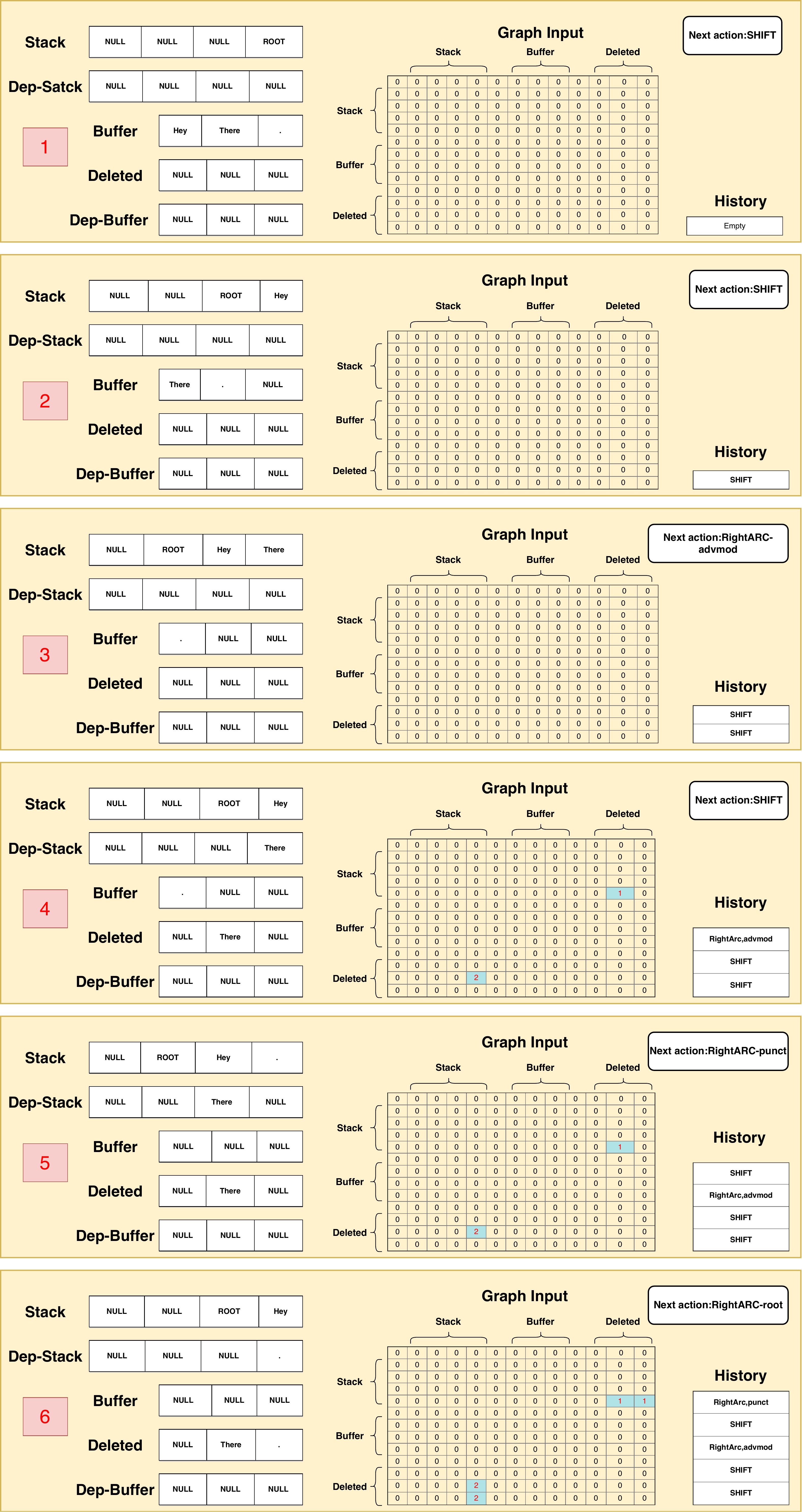}
  \caption{StateTr+G2G-Tr.}
  \label{fig:statetr}
\end{subfigure}
\caption{Example of Graph-to-Graph Transformer model integrated with SentTr and StateTr on UD English Treebank (initial sentence: "Hey There .").}
\end{figure*}

\section{Description of Treebanks}
\label{sup:treebank}

\subsection{English Penn Treebank Description}
The dataset can be found \href{https://catalog.ldc.upenn.edu/LDC99T42}{here} under LDC licence. Stanford PoS tagger and constituency converter can be downloaded from \href{https://nlp.stanford.edu/software/tagger.shtml}{here} and \href{https://nlp.stanford.edu/software/corenlp-backup-download.html}{here}, respectively. Here is the detailed information of English Penn Treebank:

\begin{table}[H]
  \centering
  \begin{adjustbox}{width=0.9\linewidth}
    \begin{tabular}{c|c|c|c|c|c} 
      Language & Version & Non-projectivity ratio & Train size(2-21) & Development size(22,24) & Test size(23) \\
      \hline
      English & 3 & 0.1\% & 39'832 & 3'046 & 2'416 \\
      \hline
    \end{tabular}
    \end{adjustbox}
    \caption{Description of English Penn Treebank.}

\end{table}

\subsection{UD Treebanks Description}

UD Treebanks v2.3 are provided in \href{https://universaldependencies.org/}{here}. Pre-processing tools can be found \href{https://universaldependencies.org/tools.html}{here}. 

\begin{table}[!ht]
  \begin{adjustbox}{width=\linewidth}
    \begin{tabular}{c|c|c|c|c|c|c|c} 
      Language & Family & Treebank & Order & Train size & Development size & Test size & Non-projectivity ratio\\
      \hline
      
      Arabic & non-IE & PADT & VSO & 6.1K & 0.9K & 0.68K & 9.2\% \\
      Basque & non-IE & BDT & SOV &  5.4K & 1.8K & 1.8K & 33.5\% \\
      Chinese  & non-IE & GSD & SVO & 4K & 0.5K & 0.5K & 0.6\% \\
      English & IE & EWT & SVO &  12.5K & 2k & 2.1K & 5.3\% \\
      Finnish & non-IE & TDT & SVO &  12.2K & 1.3K & 1.5K & 6.2\%\\
      Hebrew & non-IE & HTB & SVO & 5.2K & 0.48K & 0.49K & 7.6\%\\
      Hindi & IE & HDTB & SOV & 13.3K & 1.7K & 1.7K & 13.8\% \\
      Italian & IE & ISDT & SVO & 13.1K & 0.56K & 0.48K & 1.9\% \\
      Japanese & non-IE & GSD & SOV & 7.1K & 0.51K & 0.55K & 2.7\% \\
      Korean & non-IE & GSD & SOV & 4.4K & 0.95K & 0.99K & 16.2\% \\
      Russian & IE & SynTagRus & SVO & 48.8K & 6.5K & 6.5K & 8.0\% \\
      Swedish & IE & Talbanken & SVO & 4.3K & 0.5K & 1.2K & 3.3\% \\
      Turkish & non-IE & IMST & SOV & 3.7K & 0.97K & 0.97K & 11.1\%\\
      \hline
    \end{tabular}
    \end{adjustbox}
    \caption{Description of languages chosen from UD Treebanks v2.3.}

\end{table}

\section{Running Details of Proposed Models}
\label{sup:run-detail}
We provide the number of parameters and average run times for each model. For a better understanding, average run time is computed per transition (Second/transition). All experiments are computed with graphics processing unit (GPU), specifically the NVIDIA V100 model. The total number of transitions in the train and development sets are 79664 and 6092, respectively.

\begin{table}[H]
	\centering
	\begin{adjustbox}{width=\linewidth}
	\begin{tabular}{|l|c|c|c|c|c|}
		\hline
		Model & No. parameters & Train~(sec/transition) & Evaluation~(sec/transition) & Evaluation~(sent/sec) & Evaluation~(token/sec) \\
        \hline
		StateTr & 90.33M & 0.098 & 0.031 & 16.2 & 388.13 \\
		StateTr+G2GTr & 105.78M & 0.226 & 0.071 & 7.05 & 168.91 \\
		SentTr & 63.23M & 0.112 & 0.026 & 19.27 & 460.6 \\
		SentTr+G2GTr & 63.27M & 0.138 & 0.031 & 16.2 & 388.13 \\
		\hline
	    \end{tabular}
	\end{adjustbox}
	\caption{Running details of our models on WSJ Penn Treebank.
          \vspace{-1ex}
        } 
\end{table}

\section{Hyper-parameters for our parsing models}
\label{sup:hyper-par}
For hyper-parameter selection, we use manual tuning to find the best numbers. For BERT~\cite{bert:18} hyper-parameters, we apply the same optimization strategy as suggested by \newcite{Wolf2019HuggingFacesTS}. For classifiers and composition model, we use a one-layer feed-forward neural network for simplicity. Then, we pick hyper-parameters based on previous works~\cite{bert:18,dyer-etal-2015-transition}. We use two separate optimisers for pre-trained parameters (BERT here) and randomly initialised parameters for better convergence that is shown to be useful in ~\newcite{Kondratyuk_2019}. Early stopping (based on LAS) is used during training. The only tuning strategy that has been tried is to use one optimiser for all parameters or two different optimisers for pre-trained parameters and randomly initialised ones. For the latter case, Learning rate for randomly initialised parameters is set to $1e-4$. Results of different variations on the development set of WSJ Penn Treebank are as follows:

\begin{table}[H]
	\centering
    \begin{minipage}{0.5\linewidth}
	\begin{tabular}{|l|c|c|}
		\hline
		Model & UAS & LAS \\
        \hline
		BERT StateTr with one optimiser & 94.66 & 91.94 \\
		BERT StateTr with two optimisers & 94.24 & 91.67 \\
		Expected (Average) Performance & 94.45 & 91.81 \\
		\hline
		BERT StateTr+G2GTr with one optimiser & 94.96 & 92.88 \\
		BERT StateTr+G2GTr with two optimisers & 94.75 & 92.49 \\
		Expected (Average) Performance & 94.86 & 92.69 \\
		\hline
		BERT SentTr with one optimiser & 95.34 & 93.29 \\
		BERT SentTr with two optimisers & 95.49 & 93.29 \\
		Expected (Average) Performance & 95.42 & 93.29 \\
		\hline
		BERT SentTr+G2GTr with one optimiser & 95.27 & 93.18 \\
		BERT SentTr+G2GTr with two optimisers & 95.66 & 93.60 \\
		Expected (Average) Performance & 95.47 & 93.40 \\
		\hline
	    \end{tabular}
        \vspace{-1ex}
	\end{minipage}
	\caption{ Results on the development set of WSJ Penn Treebank for different optimisation strategy.
          \vspace{-1ex}
        } 
\end{table}

Hyper-parameters for training our models are defined as~\footnote{For UD Treebanks, we train our model for 20 epochs, and use "bert-multilingual-cased" for the initialisation. We use pre-trained BERT models of \url{https://github.com/google-research/bert}.}:

\begin{table}[H]
\centering
\begin{minipage}{.5\linewidth}
 \centering
    \begin{tabular}{c|c} 
      Component & Specification \\
      \hline
      \textbf{Optimiser} & BertAdam \\
      Learning Rate & 1e-5 \\
      Base Learning Rate & 1e-4 \\
      Adam Betas($b_1$,$b_2$) & (0.9,0.999) \\
      Adam Epsilon & 1e-6 \\
      Weight Decay & 0.01 \\
      Max-Grad-Norm & 1 \\
      Warm-up & 0.01 \\
      \hline
      \textbf{Self-Attention} \\
      No. Layers($n$) & 6 \\
      No. Heads & 12 \\
      Embedding size & 768 \\
      Max Position Embedding & 512 \\
      BERT model & bert-base-uncased\\
      \hline
      \textbf{Classifiers} \\
      No. Layers & 2 \\
      Hidden size(Exist) & 500 \\
      Hidden size(Relation) & 100 \\
      Drop-out & 0.05 \\
      Activation & $ReLU$ \\
      \hline
      \textbf{History Model} & LSTM \\
      No. Layers & 2 \\
      Hidden Size & 768 \\
      \hline
      \textbf{Composition Model} \\
      No. Layers & 2 \\
      Hidden size & 768 \\
      \hline
      Epochs & 12 \\
      \hline
    \end{tabular}
    \caption{\label{suptab:wsjhyper} Hyper-parameters for training our models.}
\end{minipage}
\end{table}

\section{Pseudo-Code of Graph Input Mechanism}
\label{sup:graph-input}

\begin{minipage}{0.46\textwidth}
\begin{algorithm}[H]
    \centering
    \begin{algorithmic}[1]
        \State{Graph Sentence(input of attention): $P$}
        \State{Graph Input: $G$}
        \State{Actions: $A = (a_1,...,a_T)$}
        \State{Input: $(S,B,D)$}
        \For{$k \gets 1, T$}
            \If{$a_k$ = \text{{\tt SHIFT} or {\tt SWAP}}}
            \State{continue}
            \Else
                \State{new relation:$i\xrightarrow{l}j$}
                \State{$G_{i,j} = 1$}
                \State{$G_{j,i} = 2$}
                \State{pop $x_j$ from stack}
                \State{change mask of $x_j$ to one}
                \State{add $l$ to input embedding of $x_j$}
                \State{ $P$:select $G$ based on Input $(S,B,D)$}
            \EndIf
        \EndFor
    \end{algorithmic}
    \caption{Pseudo-code of building graph input matrix for StateTr+G2GTr model.}
\end{algorithm}
\end{minipage}
\hfill
\begin{minipage}{0.46\textwidth}
\begin{algorithm}[H]
    \centering
    \begin{algorithmic}[1]
        \State{Graph Sentence(input of attention): $P$}
        \State{Graph Input: $G$}
        \State{Actions: $A = (a_1,...,a_T)$}
        \State{Input: initial tokens}
        \For{$k \gets 1, T$}
            \If{$a_k$ = \text{{\tt SHIFT} or {\tt SWAP}}}
            \State{continue}
            \Else
                \State{new relation:$i\xrightarrow{l}j$}
                \State{$G_{i,j} = 1$}
                \State{$G_{j,i} = 2$}
                \State{add $l$ to input embedding of $j$-th word}
                \State{$P$ = $G$}
            \EndIf
        \EndFor
    \end{algorithmic}
    \caption{Pseudo-code of building graph input matrix for SentTr+G2GTr model.}
\end{algorithm}
\end{minipage}

\section{UD Treebanks Results}
\label{suptab:scoreud-uas}
BERT SentTr+G2GTr results:

\begin{table}[htb]
	\centering
    \begin{minipage}{0.4\linewidth}
	
	\begin{adjustbox}{width=\linewidth}
	\begin{tabular}{|l|c|c|c|}
		\hline
		Language & Test set-UAS & Dev set-UAS & Dev set-LAS \\
        \hline
		Arabic & 87.65 & 87.01 & 82.64 \\
		Basque & 87.17 & 86.53 & 83.25 \\ 
		Chinese & 89.74 & 88.36 & 85.79 \\
		English & 92.05 & 93.05 & 91.3 \\
		Finnish & 91.46 & 91.37 & 89.72 \\
		Hebrew & 90.85 & 91.92 & 89.55 \\
		Hindi & 95.77 & 95.86 & 93.17 \\
		Italian & 95.15 & 95.22 & 93.9 \\
		Japanese & 96.21 & 96.68 & 96.04 \\
		Korean & 89.42 & 87.57 & 84.94 \\
		Russian & 94.42 & 94.0 & 92.70 \\
		Swedish & 92.49 & 87.26 & 85.39 \\
		Turkish & 74.23 & 72.52 & 66.05 \\
		\hline
		Average & 90.51 & 89.80 & 87.27 \\
		\hline
	    \end{tabular}
	\end{adjustbox}
        \vspace{-1ex}
	\end{minipage}
	\caption{Dependency scores of BERT SentTr+G2GTr model on the development and test sets of UD Treebanks.
          \vspace{-1ex}
        } 
\end{table}

BERT StateTr+G2GTr results:

\begin{table}[htb]
	\centering
    \begin{minipage}{0.4\linewidth}
	
	\begin{adjustbox}{width=\linewidth}
	\begin{tabular}{|l|c|c|c|}
		\hline
		Language & Test set-UAS & Dev set-UAS & Dev set-LAS \\
        \hline
		Arabic & 86.85 & 86.41 & 81.73 \\
		Basque & 80.91 & 80.01 & 73.2 \\ 
		Chinese & 87.90 & 86.64 & 84.15 \\
		English & 90.91 & 91.85 & 90.11 \\
		Finnish & 84.35 & 82.91 & 78.73 \\
		Hebrew & 89.51 & 90.36 & 87.85 \\
		Hindi & 95.65 & 95.92 & 93.30 \\
		Italian & 93.5 & 93.61 & 92.18 \\
		Japanese & 95.99 & 96.18 & 95.58 \\
		Korean & 84.35 & 82.13 & 77.78 \\
		Russian & 93.87 & 93.41 & 92.09 \\
		Swedish & 92.49 & 90.72 & 88.36 \\
		Turkish & 65.99 & 65.92 & 56.96 \\
		\hline
		Average & 87.87 & 87.39 & 84.01 \\
		\hline
	    \end{tabular}
	\end{adjustbox}
        \vspace{-1ex}
	\end{minipage}
	\caption{Dependency scores of BERT StateTr+G2GTr model on the development and test sets of UD Treebanks.
          \vspace{-1ex}
        } 
\end{table}

\section{Error-Analysis}
\label{appendix:error}
\subsection{Dependency Length}

\begin{table}[H]
  \begin{center}
    \resizebox{\textwidth}{!}{\begin{tabular}{c|c|c|c|c|c|c|c|c|c|c|c} 
      Model & ROOT & 1 & 2 & 3 & 4 & 5 & 6 & 7 & 8 & 9 & $>=$10 \\
      \hline
BERT StateTr & 96.40&94.30&93.15&91.00&87.80&85.19&82.80&81.25&80.49&82.54&84.82 \\ 
BERT StateCLSTr & 95.80&93.75&92.25&89.60&86.45&82.77&80.58&79.62&79.98&78.90&81.74 \\ 
BERT StateTr+G2GTr & 97.10&94.45&93.60&92.20&89.80&87.54&86.00&84.60&84.45&85.55&86.86 \\ 
BERT StateTr+G2CLSTr & 96.50&94.10&93.00&91.45&88.65&85.74&84.25&82.55&81.80&82.25&85.50 \\ 
BERT StateTr+G2GTr+C & 96.95&94.25&93.25&91.30&88.30&85.74&83.50&82.75&83.10&83.95&86.15 \\    
      \hline
    \end{tabular}}
    \caption{\label{suptab:sup-f1-dep} labelled F-Score vs dependency relation length}
  \end{center}
\end{table}

\begin{table}[H]
  \begin{center}
    \resizebox{\textwidth}{!}{\begin{tabular}{c|c|c|c|c|c|c|c|c|c|c|c} 
      Model & ROOT & 1 & 2 & 3 & 4 & 5 & 6 & 7 & 8 & 9 & $>=$10 \\
      \hline
BERT StateTr & 3046&31245&14478&7565&4153&2461&1681&1185&933&808&5415 \\ 
BERT StateCLSTr & 3046&31409&14473&7553&4131&2406&1641&1153&923&832&5403 \\ 
BERT StateTr+G2GTr & 3047&31240&14457&7572&4171&2465&1688&1188&941&811&5390 \\ 
BERT StateTr+G2CLSTr & 3046&31249&14447&7537&4143&2453&1701&1193&953&814&5434 \\ 
BERT StateTr+G2GTr+C & 3047&31304&14430&7514&4137&2449&1693&1182&951&830&5433 \\ 
Gold bins & 3046&31126&14490&7551&4155&2508&1698&1195&953&821&5427 \\ 
      \hline
    \end{tabular}}
    \caption{\label{suptab:sup-f1-dep-bin} Size of each bin based on dependency length}
  \end{center}
\end{table}

\subsection{Distance to Root}

\begin{table}[H]
  \begin{center}
    \resizebox{\textwidth}{!}{\begin{tabular}{c|c|c|c|c|c|c|c|c|c|c|c} 
      Model & ROOT & 1 & 2 & 3 & 4 & 5 & 6 & 7 & 8 & 9 & $>=$10 \\
      \hline
BERT StateTr & 96.40&91.35&91.00&91.45&91.80&91.50&92.10&91.75&91.90&90.06&93.06 \\ 
BERT StateCLSTr & 95.80&90.55&90.20&90.25&90.70&90.65&91.10&89.95&89.20&88.24&87.69 \\ 
BERT StateTr+G2GTr & 97.10&92.70&92.10&92.40&92.05&92.05&92.55&91.99&92.39&90.49&94.54 \\ 
BERT StateTr+G2CLSTr& 96.50&91.60&91.30&91.55&91.65&91.55&92.10&91.90&91.10&89.93&93.48 \\ 
BERT StateTr+G2GTr+C & 96.95&92.10&91.45&91.55&91.70&91.45&92.35&91.75&92.59&89.50&91.77 \\ 
      \hline
    \end{tabular}}
    \caption{\label{suptab:sup-f1-root} labelled F-Score vs distance to root}
  \end{center}
\end{table}

\begin{table}[H]
  \begin{center}
    \resizebox{\textwidth}{!}{\begin{tabular}{c|c|c|c|c|c|c|c|c|c|c|c} 
      Model & ROOT & 1 & 2 & 3 & 4 & 5 & 6 & 7 & 8 & 9 & $>=$10 \\
      \hline
BERT StateTr & 3046&16081&15965&12999&9301&6488&3964&2242&1343&740&801 \\ 
BERT StateCLSTr & 3046&15963&15798&12793&9222&6419&3974&2331&1358&827&1239 \\ 
BERT StateTr+G2GTr & 3047&16024&15889&12875&9415&6463&3995&2349&1347&743&823 \\ 
BERT StateTr+G2CLSTr & 3046&16064&15975&12971&9327&6488&3963&2279&1316&708&833 \\ 
BERT StateTr+G2GTr+C & 3047&16079&15947&13020&9419&6481&3930&2259&1274&701&813 \\ 
Gold bins & 3046&16002&16142&13064&9403&6411&3923&2298&1298&702&681 \\ 
      
      \hline
    \end{tabular}}
    \caption{\label{suptab:sup-f1-root-bin} Size of each bin based on distance to root}
  \end{center}
\end{table}

\subsection{Sentence Length}

\begin{table}[H]
  \begin{center}
    \begin{tabular}{c|c|c|c|c|c|c} 
      Model & 1-9 & 10-19 & 20-29 & 30-39 & 40-49 & $>=$ 50 \\
      \hline
      BERT StateTr & 96.1 & 95.6 & 95.0 & 94.4 & 93.9 & 90.2 \\
      BERT StateCLSTr & 94.6 & 95.0 & 94.4 & 93.9 & 93.0 & 80.2 \\
      BERT StateTr+G2GTr & 95.1 & 95.9 & 95.3 & 94.6 & 94.4 & 91.2 \\
      BERT StateTr+G2CLSTr & 94.7 & 95.1 & 94.8 & 94.2 & 93.2 & 89.4 \\
      BERT StateTr+G2GTr+C & 94.7 & 95.3 & 95.1 & 94.5 & 93.4 & 86.7 \\
      \hline
    \end{tabular}
    \caption{\label{suptab:sup-las-sent} LAS vs. sentence length}
  \end{center}
\end{table}

\subsection{Dependency Type Analysis}

\begin{table}[H]
\centering
  \begin{adjustbox}{width=0.8\linewidth}
    \begin{tabular}{|c|c|l|l|}
    \hline
    Type & StateTr+G2GTr & StateTr & StateTr+G2CLSTr \\
    \hline
{\tt acomp } & 68.62 & 58.60 (-31.9\%)  & 66.04  (-8.2\%)  \\[-0.5ex]  
{\tt advcl } & 82.75 & 70.68 (-70.0\%)  & 81.85  (-5.2\%)  \\[-0.5ex]  
{\tt advmod } & 83.85 & 84.40 (3.4\%)  & 84.35  (3.1\%)  \\[-0.5ex]  
{\tt amod } & 92.55 & 92.25 (-4.1\%)  & 92.30  (-3.4\%)  \\[-0.5ex]  
{\tt appos } & 87.85 & 84.94 (-23.9\%)  & 83.25  (-37.8\%)  \\[-0.5ex]  
{\tt aux } & 98.55 & 98.35 (-13.8\%)  & 98.45  (-6.9\%)  \\[-0.5ex]  
{\tt auxpass } & 96.65 & 96.04 (-18.0\%)  & 95.84  (-24.0\%)  \\[-0.5ex]  
{\tt cc } & 90.90 & 90.45 (-4.9\%)  & 88.80  (-23.1\%)  \\[-0.5ex]  
{\tt ccomp } & 89.49 & 81.82 (-73.0\%)  & 87.56  (-18.4\%)  \\[-0.5ex]  
{\tt conj } & 86.45 & 84.70 (-12.9\%)  & 84.15  (-17.0\%)  \\[-0.5ex]  
{\tt cop } & 93.08 & 92.62 (-6.5\%)  & 91.58  (-21.7\%)  \\[-0.5ex]  
{\tt csubj } & 76.94 & 67.93 (-39.0\%)  & 70.83  (-26.5\%)  \\[-0.5ex]  
{\tt dep } & 54.66 & 50.88 (-8.3\%)  & 51.99  (-5.9\%)  \\[-0.5ex]  
{\tt det } & 98.25 & 98.30 (2.9\%)  & 98.00  (-14.3\%)  \\[-0.5ex]  
{\tt discourse } & 15.40 & 15.40 (-0.0\%)  & 28.60  (15.6\%)  \\[-0.5ex]  
{\tt dobj } & 94.85 & 94.10 (-14.6\%)  & 93.95  (-17.5\%)  \\[-0.5ex]  
{\tt expl } & 96.39 & 94.99 (-38.8\%)  & 96.39  (-0.0\%)  \\[-0.5ex]  
{\tt infmod } & 87.38 & 79.19 (-64.9\%)  & 84.93  (-19.4\%)  \\[-0.5ex]  
{\tt iobj } & 88.01 & 90.66 (22.1\%)  & 84.24  (-31.4\%)  \\[-0.5ex]  
{\tt mark } & 95.04 & 95.19 (2.9\%)  & 94.84  (-4.1\%)  \\[-0.5ex]  
{\tt mwe } & 86.46 & 89.71 (24.0\%)  & 88.36  (14.1\%)  \\[-0.5ex]  
{\tt neg } & 95.75 & 94.84 (-21.4\%)  & 93.78  (-46.2\%)  \\[-0.5ex]  
{\tt nn } & 94.25 & 94.10 (-2.6\%)  & 93.65  (-10.5\%)  \\[-0.5ex]  
{\tt npadvmod } & 91.89 & 92.75 (10.5\%)  & 90.64  (-15.5\%)  \\[-0.5ex]  
{\tt nsubj } & 96.35 & 95.55 (-21.9\%)  & 95.60  (-20.6\%)  \\[-0.5ex]  
{\tt nsubjpass } & 95.49 & 92.70 (-61.9\%)  & 94.08  (-31.1\%)  \\[-0.5ex]  
{\tt num } & 95.25 & 94.89 (-7.4\%)  & 95.15  (-2.0\%)  \\[-0.5ex]  
{\tt number } & 92.50 & 90.65 (-24.6\%)  & 92.10  (-5.3\%)  \\[-0.5ex]  
{\tt parataxis } & 69.59 & 62.89 (-22.1\%)  & 72.10  (8.2\%)  \\[-0.5ex]  
{\tt partmod } & 82.11 & 72.82 (-52.0\%)  & 79.74  (-13.3\%)  \\[-0.5ex]  
{\tt pcomp } & 88.12 & 86.49 (-13.7\%)  & 85.77  (-19.8\%)  \\[-0.5ex]  
{\tt pobj } & 97.15 & 96.95 (-7.0\%)  & 96.60  (-19.3\%)  \\[-0.5ex]  
{\tt poss } & 97.60 & 97.15 (-18.7\%)  & 97.70  (4.2\%)  \\[-0.5ex]  
{\tt possessive } & 98.29 & 98.04 (-14.5\%)  & 98.44  (8.9\%)  \\[-0.5ex]  
{\tt preconj } & 85.71 & 84.65 (-7.5\%)  & 84.65  (-7.5\%)  \\[-0.5ex]  
{\tt predet } & 79.34 & 79.34 (-0.0\%)  & 77.44  (-9.2\%)  \\[-0.5ex]  
{\tt prep } & 90.25 & 89.90 (-3.6\%)  & 89.50  (-7.7\%)  \\[-0.5ex]  
{\tt prt } & 84.48 & 83.42 (-6.9\%)  & 83.10  (-8.9\%)  \\[-0.5ex]  
{\tt punct } & 88.45 & 88.05 (-3.5\%)  & 87.65  (-6.9\%)  \\[-0.5ex]  
{\tt quantmod } & 86.97 & 84.40 (-19.7\%)  & 84.49  (-19.0\%)  \\[-0.5ex]  
{\tt rcmod } & 86.84 & 76.38 (-79.5\%)  & 83.91  (-22.3\%)  \\[-0.5ex]  
{\tt root } & 97.15 & 96.40 (-26.3\%)  & 96.50  (-22.8\%)  \\[-0.5ex]  
{\tt tmod } & 86.02 & 86.38 (2.6\%)  & 85.03  (-7.1\%)  \\[-0.5ex]  
{\tt xcomp } & 90.75 & 84.44 (-68.2\%)  & 89.75  (-10.8\%)  \\[-0.5ex]  
    \hline
    \end{tabular}
  \end{adjustbox}
\caption{F-score of StateTr, StateTr+G2GTr, and StateTr+G2CLSTr models for the dependency types on the development set of WSJ Treebank. Relative error reduction is computed by considering StateTr+G2GTr scores as the reference.}
\label{sup:acc-deptype}
\end{table}

\end{appendices}

\end{document}